\pdfoutput=1
\documentclass[11pt]{article}

\usepackage[final]{acl}

\usepackage{float}
\usepackage{times}
\usepackage{latexsym}

\usepackage[T1]{fontenc}
\usepackage[utf8]{inputenc}

\usepackage{microtype}

\usepackage{inconsolata}

\usepackage{graphicx}
\usepackage{booktabs}
\usepackage{longtable}
\usepackage{amsmath}
\usepackage{natbib}

\title{Rice-VL: Evaluating Vision-Language Models for Cultural Understanding Across ASEAN Countries}

\author{
Tushar Pranav\textsuperscript{1}$^*$,
Eshan Pandey\textsuperscript{1}\thanks{Equal contribution.},
Austria Lyka Diane Bala\textsuperscript{1},
Aman Chadha\textsuperscript{2}\thanks{Work done outside role at Amazon.}, \\
Indriyati Atmosukarto\textsuperscript{1},
Donny Soh Cheng Lock\textsuperscript{1} \\
\textsuperscript{1}Singapore Institute of Technology \\
\textsuperscript{2}Amazon GenAI, Palo Alto, CA, USA \\
\texttt{\{pranav.tushar, pandey.eshan, lyka.austria, indriyati, donny.soh\}@singaporetech.edu.sg}, \\
\texttt{hi@aman.ai}
}

\begin{document}
\maketitle
% \begin{abstra
\begin{abstract}
Vision-Language Models (VLMs) excel in multimodal tasks but often exhibit Western-centric biases, limiting their effectiveness in culturally diverse regions like Southeast Asia (SEA). To address this, we introduce RICE-VL, a novel benchmark evaluating VLM cultural understanding across 11 ASEAN countries. RICE-VL includes over 28,000 human-curated Visual Question Answering (VQA) samples—covering True/False, Fill-in-the-Blank, and open-ended formats—and 1,000 image-bounding box pairs for Visual Grounding, annotated by culturally informed experts across 14 sub-ground categories. We propose SEA-LAVE, an extension of the LAVE metric, assessing textual accuracy, cultural alignment, and country identification. Evaluations of six open- and closed-source VLMs reveal significant performance gaps in low-resource countries and abstract cultural domains. The Visual Grounding task tests models’ ability to localize culturally significant elements in complex scenes, probing spatial and contextual accuracy. RICE-VL exposes limitations in VLMs’ cultural comprehension and highlights the need for inclusive model development to better serve diverse global populations.
\end{abstract}

\section{Introduction}

% \section{Introduction}
\label{sec:intro}

\begin{figure}
  \centering
    \includegraphics[width=0.98\linewidth]{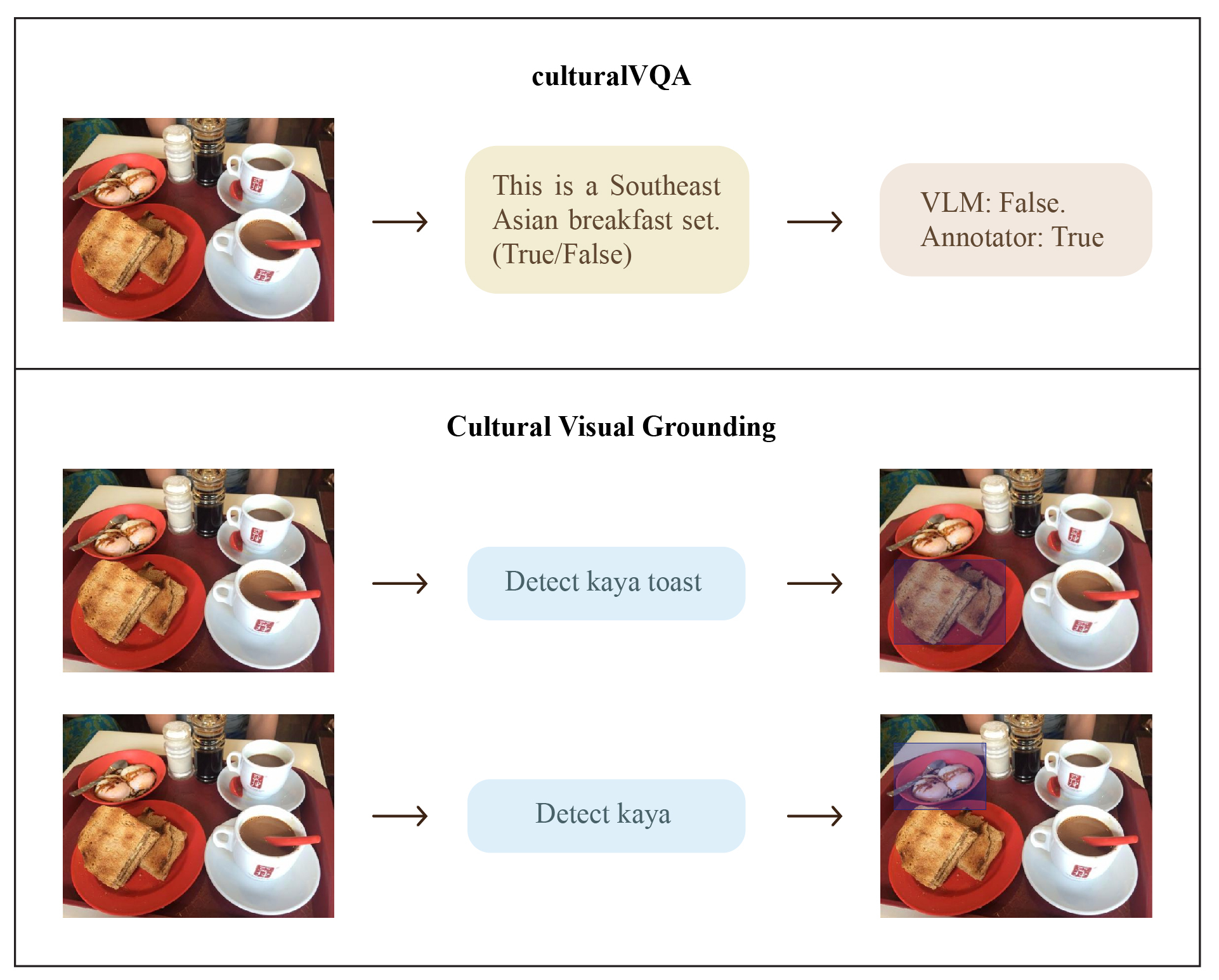}
  \caption{An example instance from each task in RICE-VL Benchmark: i)  culturalVQA; ii) cultural Visual Grounding.}
  \label{fig:intro}
\end{figure}

\begin{table*}[ht]
\centering
\resizebox{\textwidth}{!}{ 
\begin{tabular}{l l l l}
\hline
\textbf{Feature} & \textbf{SEA-Crowd} & \textbf{SEA-VL} & \textbf{RICE-VL} \\ 
\hline
\textbf{Primary Focus} & Multilingual and multimodal data aggregation & Large-scale culturally relevant image dataset & Cultural reasoning and evaluation in VLMs \\
% \hline
\textbf{Data Modalities} & Text, audio, image & Image & Image with culturally annotated tasks \\ 
% \hline
\textbf{Data Collection Methods} & Aggregation of existing datasets & Crowdsourcing, web crawling, synthetic generation & Curated by trained cultural annotators \\
% \hline
\textbf{Evaluation Tasks} & Language processing tasks across modalities & Image captioning, retrieval & VQA (including True/False, Fill in the Blanks), VG \\ 
% \hline
\textbf{Cultural Reasoning Emphasis} & Limited & Moderate & High \\ 
% \hline
\textbf{Human-Centric Annotations} & Yes & Partially (crowdsourced and synthetic) & Yes (trained cultural experts) \\
\hline
\end{tabular}
}
\caption{\label{tab:comparison} Comparative analysis of SEA-Crowd, SEA-VL, and RICE-VL benchmarks.}
\end{table*}

The advancement of large vision-language models (LVLMs) \cite{achiam2023gpt,bai2023qwen,beyer2024paligemma,liu2023visual} has propelled substantial progress in multimodal tasks such as image captioning, visual question answering, and dialogue generation. However, a critical gap persists in their ability to effectively interpret and respond to culturally specific concepts, particularly within diverse and low-resource regions like Southeast Asia \cite{aji2022one,yong2023prompting, myung2024blend}. While existing LVLMs demonstrate robust performance on datasets grounded in high-resource, Western-centric contexts, they often struggle to generalize to the complex cultural nuances, hybrid traditions, and multilingual environments characteristic of ASEAN countries \cite{cahyawijaya2025crowdsource,romero2024cvqa,liu2021visually,gustafson2023facet, shankar2017no}.  Current benchmarks assessing cultural and multilingual competence in vision-language models are predominantly focused on Western and  Anglocentric settings, resulting in a significant underrepresentation of cultural richness from countries such as Indonesia, Vietnam, the Philippines, and Myanmar—regions distinguished by unique visual and linguistic markers shaped by centuries of local tradition, colonial influence, and contemporary globalization This Western-centric bias underscores the pressing need for culturally diverse benchmarks to systematically evaluate and enhance the cultural inclusiveness and alignment of modern LVLMs. 

In response to these challenges, we propose \textbf{RICE-VL}, a comprehensive benchmark explicitly designed to evaluate the cultural understanding and contextual reasoning capabilities of VLMs in Southeast Asia. The RICE-VL benchmark consists of two core tasks: culturalVQA and cultural visual grounding, adapted from prior works, including culturalVQA \cite{nayak2024benchmarking} and cultural visual grounding from the globalRG benchmark \cite{bhatia2024local}. Figure \ref{fig:intro} presents the examples of these two tasks.

The \textbf{ culturalVQA} task consists of three core components: Question Answering, True or False, and Fill in the Blanks, requiring models to integrate both visual and textual information. This structure provides a comprehensive framework for evaluating the models' capability to recognize and reason about cultural nuances across diverse contexts.

The \textbf{cultural visual grounding }task requires models to pinpoint specific coordinates of culturally relevant elements depicted in the images, assessing their spatial understanding of cultural representations. 

Our evaluation on state-of-the-art VLMs reveals a persistent gap in cultural understanding, particularly concerning low-resource Southeast Asian cultures. Closed-source models such as GPT-4O and Claude-3-Opus outperform open-source counterparts across most countries, but all models demonstrate reduced accuracy in underrepresented regions like Timor-Leste, Brunei, and Laos. 

Our contributions are as follows:

• We present RICE-VL, a culturally diverse multimodal benchmark designed to capture the rich cultural context of ASEAN countries. Comprising over 28,000 question-answer tasks for VQA based on 7000 images, and 1,000 image-bounding box tasks for Visual Grounding, the benchmark offers extensive coverage across 11 ASEAN countries, encompassing a comprehensive range of cultural themes.

• The dataset is systematically developed and rigorously validated by annotators trained in cultural contexts over a comprehensive 720-hour annotation period (6 annotators, 6 hours/day, 20 days), ensuring cultural relevance and accuracy across both low- and high-resource ASEAN countries.

• We benchmark existing state-of-the-art VLMs on RICE-VL, identifying key performance gaps and areas for improvement, with particular attention to the influence of Western centric biases on model performance.

\section{Related Works}

\begin{table*}
    \centering
    \small % Reduces font size
    \setlength{\tabcolsep}{5pt} % Adjust column spacing
    \renewcommand{\arraystretch}{1.1} % Adjust row height
    \resizebox{\textwidth}{!}{ % Automatically resizes table to fit page width
    \begin{tabular}{l c c c c c c}
        \toprule
         \hline
        Country & Claude-3-Opus & GPT-4O & LLaMA 3.2 (11B) & Ola (7B) & Ovis 2 (8B) & Qwen-VL 2.5 (7B) \\
        \hline
        \midrule
        Brunei        & 0.58 & 0.55 & 0.50 & 0.33 & 0.53 & 0.44 \\
        Cambodia      & 0.66 & 0.64 & 0.62 & 0.45 & 0.52 & 0.53 \\
        Indonesia     & 0.73 & 0.74 & 0.66 & 0.54 & 0.64 & 0.62 \\
        Laos          & 0.54 & 0.52 & 0.38 & 0.29 & 0.46 & 0.38 \\
        Malaysia      & 0.78 & 0.77 & 0.74 & 0.54 & 0.66 & 0.58 \\
        Myanmar       & 0.60 & 0.54 & 0.50 & 0.45 & 0.51 & 0.49 \\
        Philippines   & 0.67 & 0.65 & 0.63 & 0.49 & 0.50 & 0.43 \\
        Singapore     & 0.79 & 0.73 & 0.70 & 0.43 & 0.64 & 0.57 \\
        Thailand      & 0.82 & 0.72 & 0.65 & 0.59 & 0.63 & 0.64 \\
        Timor-Leste   & 0.40 & 0.26 & 0.21 & 0.17 & 0.19 & 0.20 \\
        Vietnam       & 0.63 & 0.59 & 0.48 & 0.45 & 0.54 & 0.42 \\
         \hline
        \bottomrule
    \end{tabular}
    }
    \caption{SEA-LAVE scores for various open-source and closed-source models on CulturalVQA task.}
    \label{tab:model_country_scores}
\end{table*}

The development of large vision-language models (VLMs) has significantly advanced multimodal tasks, yet their performance often reflects a Western-centric bias due to the predominance of Anglocentric datasets like MSCOCO \cite{lin2014microsoft} and Visual Genome \cite{krishna2017visual}. These datasets primarily feature imagery and contexts from Western cultures, limiting VLMs’ ability to generalize to culturally diverse regions in Southeast Asia (SEA), which encompasses over 1,300 languages and 11 countries \cite{cahyawijaya2025crowdsource}. This bias underscores the need for culturally inclusive benchmarks to evaluate VLMs’ understanding of non-Western cultural nuances.

Recent initiatives have increasingly sought to address the regional imbalances in AI benchmarks and datasets, particularly focusing on culturally nuanced multimodal resources. Community-driven efforts such as AI4Bharat \cite{nath2025vision1language}, Sarvam AI \cite{khan2024indicllmsuite}, and Krutruim AI \cite{khan2025chitrarth} have laid foundational work in Indic AI, developing benchmarks and datasets that encapsulate regional linguistic and cultural contexts. Similarly, Chinese AI labs have advanced the development of culturally specific benchmarks, as evidenced by initiatives like CVLUE \cite{wang2025cvlue}, VisTW: \cite{tam2025vistw} and associated datasets.

In Southeast Asia, emerging benchmarks like ViOCRVQA \cite{pham2025viocrvqa}
and MalayMMLU \cite{poh-etal-2024-malaymmlu} are contributing to the landscape by introducing visual and language tasks for Vietnamese and Malay, respectively. However, these efforts remain relatively isolated, underscoring the persistent need for cohesive, culturally diverse benchmarks that not only capture regional nuances but also facilitate robust evaluation of multimodal AI systems in Southeast Asia.however these are llimitted to individual countrues and there is a need for a collective community led invitiate for preserving the cultural and local values and aspects of sea countries?  
Notable among these is SEA-Crowd \cite{lovenia2024seacrowd}, and Sea-VL, which aggregates data spanning text, audio, and images for nearly 1,000 SEA languages, encompassing 13 tasks and 36 indigenous languages. However, SEA-Crowd's primary emphasis remains on language processing, with limited exploration of visual-cultural reasoning.
 
In Southeast Asia, SEA-Crowd \cite{lovenia2024seacrowd} aggregates multimodal resources across text, audio, and images for nearly 1,000 SEA languages, supporting 13 tasks and 36 indigenous languages. However, its focus remains on language processing, with limited emphasis on visual-cultural tasks . SEA-VL \cite{cahyawijaya2025crowdsource} compiles 1.28 million culturally relevant images through crowdsourcing, web crawling, and
synthetic generation, but its evaluation centers on descriptive tasks like image captioning and retrieval, which do not fully capture the depth of cultural understanding required for SEA contexts. Table~\ref{tab:comparison} provides a comparative study of three major Southeast Asian benchmarks—SEA-Crowd, SEA-VL, and RICE-VL—highlighting their differences in focus, modalities, data collection methods, and emphasis on cultural reasoning.

In contrast, RICE-VL is purpose built to evaluate VLMs’ cultural understanding and contextual reasoning in Southeast Asia. It comprises two tasks: culturalVQA, covering question answering, true/false, and fill-in-the-blanks; and cultural visual grounding, which assesses the localization of culturally salient elements. Unlike SEA-VL’s reliance on synthetic data, RICE-VL uses 720 hours of expert human annotation to ensure cultural accuracy and depth. RICE-VL goes beyond surface-level evaluations by focusing on culturally grounded reasoning and localization tasks. It highlights significant performance gaps in existing VLMs—especially in low-resource countries—and calls for benchmarks that emphasize cultural alignment, not just data diversity.

\section{Task 1: Cultural Visual Question Answering}

\begin{figure*}[t]
  \includegraphics[width=0.48\linewidth]{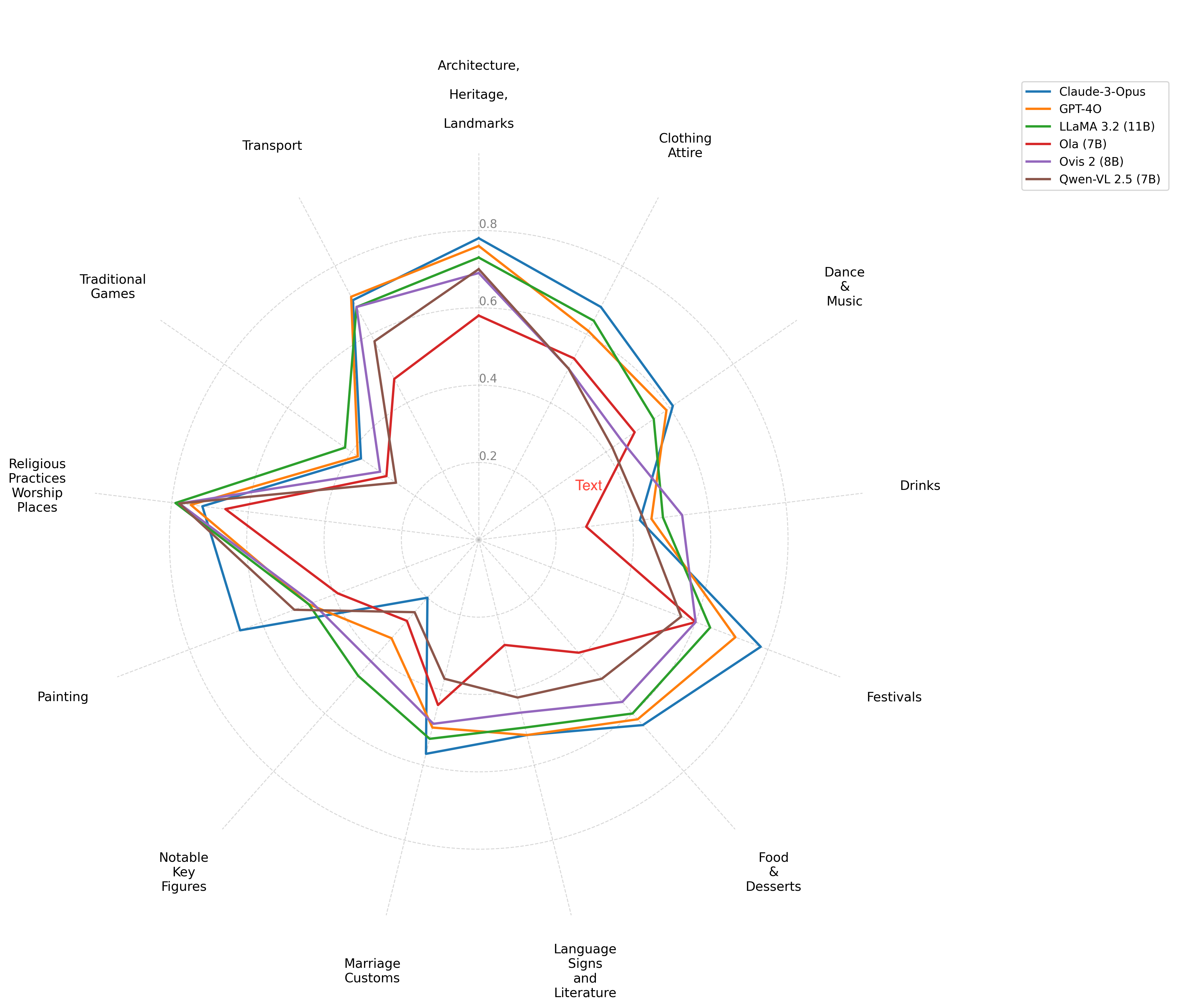} \hfill
  \includegraphics[width=0.48\linewidth]{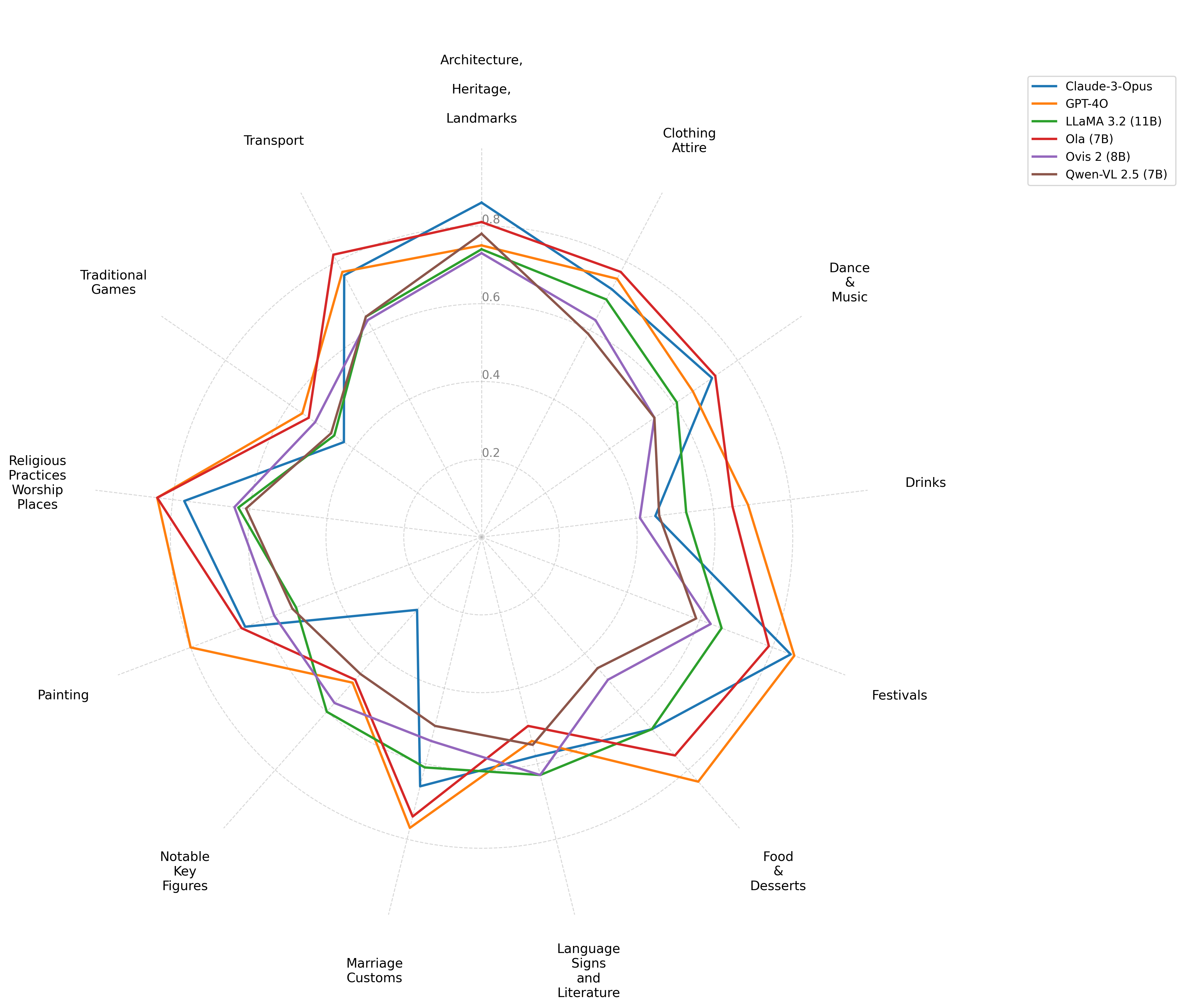}
  \caption {Cultural understanding of various models assessed on culturalVQA tasks, when the model was prompted global context (left) and with SEA specific context (right)}
   \label{fig:task1_figs}
\end{figure*}

\begin{table*}
    \centering
    \small
    \setlength{\tabcolsep}{5pt}
    \renewcommand{\arraystretch}{1.1}
    \resizebox{\textwidth}{!}{
    \begin{tabular}{l c c c c c c}
        \toprule
        \hline
        Country & Claude-3-Opus & GPT-4O & LLaMA 3.2 (11B) & Ola (7B) & Ovis 2 (8B) & Qwen-VL 2.5 (7B) \\
        \hline
        \midrule
        Brunei        & 0.63 & 0.61 & 0.50 & 0.63 & 0.44 & 0.42 \\
        Cambodia      & 0.76 & 0.70 & 0.58 & 0.68 & 0.65 & 0.51 \\
        Indonesia     & 0.80 & 0.79 & 0.72 & 0.80 & 0.78 & 0.81 \\
        Laos          & 0.64 & 0.60 & 0.39 & 0.59 & 0.56 & 0.53 \\
        Malaysia      & 0.75 & 0.80 & 0.77 & 0.82 & 0.71 & 0.79 \\
        Myanmar       & 0.70 & 0.65 & 0.53 & 0.72 & 0.42 & 0.36 \\
        Philippines   & 0.59 & 0.85 & 0.67 & 0.71 & 0.53 & 0.43 \\
        Singapore     & 0.75 & 0.87 & 0.72 & 0.72 & 0.67 & 0.75 \\
        Thailand      & 0.78 & 0.49 & 0.72 & 0.87 & 0.71 & 0.68 \\
        Timor-Leste   & 0.34 & 0.71 & 0.22 & 0.46 & 0.18 & 0.25 \\
        Vietnam       & 0.59 & 0.63 & 0.58 & 0.66 & 0.74 & 0.59 \\
        \hline
        \bottomrule
    \end{tabular}
    }
    \caption{SEA-LAVE scores for various open-source and closed-source models on CulturalVQA task with cultural context in the prompt.}
    \label{tab:model_country_scores_improved}
\end{table*}

\subsection{Data Collection, Annotation, and Verification}

\paragraph{Data Collection .}
Data collection for the culturalVQA task was carried out in 11 Southeast Asian countries, encompassing Singapore, Malaysia, Timor-Leste, Vietnam, the Philippines, Indonesia, Brunei, Laos, Myanmar, Thailand, and Cambodia. The dataset was stratified into cultural domains such as Architecture and Heritage, Clothing and Attire, Dance and Music, Drinks, Festivals, Food and Desserts, Language Signs and Literature, Marriage Customs, Notable Key Figures, Painting, Religious Practices, Places of Worship, Traditional Games, and Transport. Each domain was further divided into 10 subcultures, each represented by 5 to 25 images. For instance, in the 'Food and Desserts' category for Singapore, subcultures include Bak Kut Teh, Rojak, Char Kway Teow, Chendol, Chilli Crab, and Hainanese Chicken Rice. Data acquisition employed web scraping targeting culturally specific visual content across these subcategories.

\paragraph{Annotation .}
The annotation process for culturalVQA was structured to capture cultural nuances and ensure accurate visual representation. Annotators, specifically trained in identifying culturally significant elements, curated visual question-answer pairs. Each image was assigned two initial questions generated using GPT-4.0 \cite{achiam2023gpt} based on metadata, followed by five additional questions curated by annotators. The questions encompassed True/False and Fill-in-the-Blanks formats, emphasizing cultural-contextual reasoning.

\paragraph{Verification .}
Verification procedures for culturalVQA involved multiple rounds of cultural relevance checks. Annotators from Southeast Asia reviewed each image-question pair to confirm cultural accuracy and prevent content bias. Discrepancies identified during verification were addressed through iterative reviews, ensuring that each visual question-answer pair effectively conveyed the intended cultural concept without introducing ambiguity.

\subsection{Task Definition and Evaluation Setup}

The culturalVQA task is designed to evaluate a model’s ability to accurately interpret and reason about culturally specific visual content within the context of Southeast Asian cultural domains. Given an image and a corresponding question, the model is expected to generate culturally appropriate responses that reflect the visual content while aligning with the cultural context depicted. The task encompasses various question formats, including True/False, Fill-in-the-Blanks, and open-ended questions, enabling a comprehensive assessment of the model’s cultural reasoning capabilities across multiple dimensions. 

Additionally, all experiments  are conducted under two distinct settings: a Global setting and a Southeast Asian specific setting. In the Global setting, the prompt includes the instruction “This is a global setting”, followed by questions  encouraging open-ended, worldwide reasoning. In contrast, the SEA specific setting includes the instruction “This is a Southeast Asian setting” to anchor the model within a regional context. This dual-setting design allows us to systematically evaluate how prompt framing influences the model’s cultural localization performance and whether regional cues enhance its ability to reason about culturally specific content.

Evaluating cultural reasoning in multimodal tasks presents distinct challenges, particularly when scaling assessments across culturally diverse datasets. Traditional evaluation frameworks primarily rely on string matching techniques to measure alignment between model-generated outputs and ground-truth data \cite{nayak2024benchmarking}. However, recent studies highlight the potential of using large language models (LLMs) as evaluators, acting as adjudicators to assess the contextual and cultural accuracy of responses \cite{manas2024improving, nayak2024benchmarking}.

Building on these insights, we introduce the Southeast Asia Linguistic Agreement with Visual Evidence (SEA-LAVE) metric, an adaptation of the original LAVE metric \cite{manas2024improving} that incorporates a cultural dimension to better align with the objectives of our benchmark. For evaluation, we employ the Qwen2.5-VL 7B model as the reference LLM, given its strong open-source performance and reproducibility. We deliberately exclude proprietary models such as GPT-4 and Claude from the evaluation step to ensure transparency and replicability of results.

\paragraph{SEA-LAVE Metric.}
To address this, we adopt and extend the Linguistic Agreement with Visual Evidence (LAVE) metric by introducing a culturally grounded variant: \textbf{SEA-LAVE}. This metric assesses alignment between the model’s response and expected output across three dimensions: textual relevance, cultural appropriateness, and regional specificity.

% \begin{figuret]
\begin{equation}
\textbf{SEA-LAVE} = \frac{\text{TU} + \text{CU} + \left(\frac{\text{CI}}{2}\right)}{3}
\end{equation}
% \end{figure*}

\noindent Each component is a binary score (0 or 1), determined through either human annotation or LLM-based evaluation:
\begin{itemize}
\item \textbf{Text Understanding (TU):} Assesses semantic alignment between the model output and expected answer.
\item \textbf{Cultural Understanding (CU):} Evaluates the response’s adherence to the relevant cultural context or practice.
\item \textbf{Country Identification (CI):} Measures whether the model correctly identifies the Southeast Asian country, with partial weighting to account for ambiguity across borders.
\end{itemize}

By incorporating cultural specificity into the scoring, SEA-LAVE offers a more holistic metric for benchmarking cross-cultural competence in vision-language models, particularly within the diverse sociocultural landscape of Southeast Asia.

\subsection{Models}

For the VQA task, we benchmarked six VLMS selected based on their applicability to cultural visual reasoning and their performance in VLM leaderboards \cite{duan2024vlmevalkit}. The models are categorized into four open-source and two closed-source systems. The open-source models include Qwen-VL 2.5 (7B) \cite{bai2023qwen}, Ovis 2 (8B) \cite{lu2024ovis}, LLaMA 3.2 (11B) \cite{grattafiori2024llama}, and Ola (7B) \cite{liu2025ola}. The closed-source models consist of GPT-4O \cite{achiam2023gpt} and Claude-3-Opus. This selection spans a range of architectures and parameter sizes, facilitating a comprehensive evaluation of cultural reasoning capabilities across both open and closed-source frameworks.

\subsection{Results and Analysis}

As illustrated in Figure~\ref{fig:task1_figs}, model responses to culturalVQA tasks show improved cultural grounding when Southeast Asian context is explicitly included in the prompt. Table~\ref{tab:model_country_scores} presents SEA-LAVE scores under the global setting (without geographic cues), while Table~\ref{tab:model_country_scores_improved} shows the corresponding scores under the SEA-specific setting. Together, these results highlight the importance of regional grounding for accurate cultural understanding across the 11 SEA countries.

\paragraph{\textbf{Do closed-source VLMs exhibit stronger cultural reasoning than open-source models?}}

Closed-source models—\textbf{Claude-3-Opus} and \textbf{GPT-4O}—consistently outperform their open-source counterparts across nearly all countries. Claude-3-Opus yields the highest SEA-LAVE scores in high-resource countries such as Malaysia, Thailand, and Indonesia, while GPT-4O demonstrates notable strength in the Philippines and Timor-Leste. 

\paragraph{\textbf{Do region-specifc models demonstrate advantages in Southeast Asian settings?}}

While region-specific models like \textbf{Qwen-VL 2.5} and \textbf{Ovis 2} exhibit improved performance in culturally diverse settings, particularly Malaysia, Indonesia, and Vietnam, they fall short of matching closed-source models in both breadth and depth of reasoning. Their strength appears to correlate with countries that have relatively higher online representation in training corpora, but their performance degrades in underrepresented contexts like Brunei and Timor-Leste. This suggests that region alone is insufficient without culturally grounded training data.

\paragraph{\textbf{Are VLMs equally capable across all SEA countries?}}

Performance varies significantly across countries, with high-resource nations (e.g., Singapore, Malaysia) yielding higher SEA-LAVE scores, and low-resource ones (e.g., Timor-Leste, Brunei, Laos) consistently underperforming across all models. Notably, Timor-Leste remains the most challenging for all systems, indicating limited representation in pretraining corpora and a lack of cultural exposure. 

\paragraph{\textbf{Does prompt framing influence cultural reasoning in VLMs?}}
We observe a marked performance boost when prompts are regionally anchored. When explicitly framed with “This is a Southeast Asian setting,” models—especially GPT-4O and Ola (7B)—show improved cultural localization. This finding affirms the importance of contextual priming in VLM prompting and suggests a simple, low-resource intervention to enhance model sensitivity to cultural cues. For instance, Ola's score on Thailand jumps from 0.59 to 0.87 with the SEA-specific prompt, an improvement not observed in the global setting.

\textbf{Collectively, these findings validate the need for culturally aware benchmarks like RICE-VL and affirm that improving cultural competence in VLMs requires both diverse training data and region-sensitive evaluation protocols.
}

\section{Task 2: Cultural Visual Grounding}

\begin{table*}[!htbp]
\centering
\resizebox{\textwidth}{!}{%
\begin{tabular}{lcccccc}
\hline
\textbf{Category} & \textbf{Paligemma2 3B} & \textbf{Paligemma2 10B} & \textbf{Qwen2.5 VL 3B} & \textbf{Qwen2.5 VL 7B} & \textbf{Kosmos2} & \textbf{GroundingDino} \\
\hline
Brunei & 0.345 & 0.408 & 0.546 & 0.531 & 0.421 & 0.380 \\
Cambodia & 0.172 & 0.222 & 0.317 & 0.312 & 0.264 & 0.271 \\
Indonesia & 0.360 & 0.520 & 0.551 & 0.494 & 0.523 & 0.452 \\
Laos & 0.433 & 0.434 & 0.535 & 0.506 & 0.449 & 0.488 \\
Malaysia & 0.355 & 0.475 & 0.548 & 0.510 & 0.492 & 0.438 \\
Myanmar & 0.395 & 0.393 & 0.458 & 0.412 & 0.369 & 0.469 \\
Philippines & 0.286 & 0.463 & 0.478 & 0.520 & 0.451 & 0.389 \\
Singapore & 0.349 & 0.454 & 0.555 & 0.527 & 0.349 & 0.427 \\
Thailand & 0.394 & 0.482 & 0.497 & 0.498 & 0.411 & 0.531 \\
Timor Leste & 0.334 & 0.438 & 0.420 & 0.428 & 0.328 & 0.417 \\
Vietnam & 0.295 & 0.282 & 0.440 & 0.390 & 0.327 & 0.343 \\
\hline
\end{tabular}}
\caption{Average IoU scores for Cultural Grounding Task across ASEAN countries using various VL models.}
\label{tab:cultural-grounding}
\end{table*}

\subsection{Data Collection, Annotation, and Verification}

\paragraph{Data Collection . } Data collection for the Visual Grounding (VG) task was systematically conducted across 11 Southeast Asian countries. The dataset was structured to represent 95 distinct cultural subcategories, including ceremonial clothing, traditional dance forms, and religious artifacts. Images were sourced using web scraping, targeting culturally significant visual content across each subcategory.

\paragraph{Annotation .} Following data collection, the annotation process focused on cultural specificity and visual clarity. Annotators underwent targeted training to identify and demarcate cultural elements amidst potential visual distractions. CVAT software was employed to annotate bounding boxes around cultural markers, resulting in 990 image-bounding box pairs. Each image was annotated to include multiple cultural markers, thereby enhancing the dataset’s complexity for cultural grounding tasks.

\paragraph{Verification .} To ensure cultural accuracy and mitigate biases, the verification process involved multiple stages of review by annotators familiar with Southeast Asian cultural contexts. Annotators validated the cultural relevance of each bounding box annotation, confirming that each visual marker accurately reflected its designated cultural category. Additionally, data integrity checks were performed to identify and rectify inconsistencies, ensuring the dataset’s robustness for downstream evaluation.

\subsection{Model}
The model selection was driven by two primary objectives: assessing grounding precision and evaluating cross-cultural alignment. Grounding Dino \cite{liu2024grounding} was selected for its targeted training on grounding-specific tasks, as demonstrated by its application in GlobalRG \cite{bhatia2024local}. Meanwhile, Qwen2.5 VL (3B and 7B) \cite{bai2023qwen} and Paligemma2 (3B and 10B) \cite{beyer2024paligemma} were included as general-purpose models, leveraging their robust visual-text alignment capabilities. Kosmos2 \cite{peng2023kosmos} was incorporated to evaluate its cross-modal grounding effectiveness across culturally diverse contexts, aligning with the comparative framework in GlobalRG \cite{bhatia2024local} to assess the performance gap between task-specific and general-purpose models.

\subsection{Task Definition and Evaluation}

Cultural visual grounding refers to the model's capability to accurately identify and localize culturally significant elements within a given image using bounding boxes. This task assesses the model’s ability to discern and demarcate cultural markers based on textual prompts, reflecting both grounding precision and cultural understanding.

Given an image \( I \) and a text prompt \( p \), the model predicts a bounding box \( \hat{B} \) corresponding to the region within \( I \) that aligns with the prompt \( p \). The ground truth bounding box is denoted as \( B \).

\textbf{Intersection over Union (IoU)} is a widely adopted metric for evaluating the overlap between predicted and ground truth bounding boxes. The IoU score quantifies the extent of overlap and is instrumental in assessing the model's grounding accuracy and cultural precision.

The IoU is calculated as:

\begin{equation}
IoU = \frac{|R_{pred} \cap R_{gtruth}|}{|R_{pred} \cup R_{gtruth}|}
\end{equation}

where:

\begin{itemize}
    \item \( R_{pred} \) denotes the predicted bounding box.
    \item \( R_{gtruth} \) represents the ground truth bounding box.
    \item \( |R_{pred} \cap R_{gtruth}| \) is the area of overlap between the predicted and ground truth bounding boxes.
    \item \( |R_{pred} \cup R_{gtruth}| \) is the total area covered by both bounding boxes.
\end{itemize}

In addition to evaluating model predictions, IoU is also employed to assess consistency among human annotators. This is achieved by comparing the IoU scores between bounding boxes drawn by multiple annotators, providing insights into cultural ambiguities and the reliability of cultural representations across different annotators.

\subsection{Results and Analysis}

Table~\ref{tab:cultural-grounding} presents the average IoU scores for the Cultural Grounding task across ASEAN countries.

\paragraph{\textbf{Can VLMs ground culturally specific markers across SEA countries with high accuracy?}}

As shown in Table~\ref{tab:cultural-grounding}, Qwen2.5-VL consistently achieves the highest average IoU scores across most Southeast Asian countries, with particularly strong performance in Singapore, Brunei, and Indonesia. These findings highlight the importance of multimodal pretraining that incorporates culturally rich image-text pairs. Models that rely predominantly on geometric alignment or generic object detection tend to underperform in contexts requiring nuanced cultural understanding. The results underscore that grounding culturally specific markers extends beyond spatial accuracy—it demands culturally aware representation learning.

\paragraph{\textbf{Can VLMs localize Southeast Asian cultural artifacts with distinct visual identity?}}

Models like Qwen2.5-VL and Paligemma2 excel at localizing culturally unique artifacts such as batik patterns (Indonesia, Malaysia) and chada headgear (Thailand). However, when visual features resemble common global objects—like kaya toast looking like Western bread—models tend to make generic predictions. This highlights a challenge in cross-cultural disambiguation, where visual similarity can overshadow cultural context. Accurate grounding thus requires both visual recognition and culturally informed understanding.

\paragraph{\textbf{Can VLMs achieve consistent grounding across different cultural categories?}}

Grounding accuracy differs significantly across the 14 cultural subdomains. Categories with clear, prominent visuals like Clothing, Transport, and Festivals achieve higher IoU scores, likely due to the size and visibility of objects. In contrast, areas involving smaller or more abstract elements—such as Religious Practices, Key Figures, and Painting—show lower accuracy across models. This gap is often due to visual clutter or symbolic imagery that makes grounding more difficult. These results highlight how category complexity and visual ambiguity impact model performance, especially in less represented cultural themes.

\section{Limitations}

While RICE-VL provides a broad assessment of cultural understanding in Vision-Language Models across ASEAN countries, it has several limitations. Due to resource constraints, our evaluation is limited to large-scale models (up to 12B parameters), leaving the performance of smaller or low-resource models largely unexplored. Future work should consider these models and techniques like distillation. Additionally, our current task formats—primarily culturalVQA and grounding—focus on visual-text alignment and may not capture deeper cultural reasoning such as historical or narrative context. More expressive tasks are needed. Lastly, the benchmark is English-only, which simplifies evaluation but may overlook culturally nuanced meanings in native languages. Incorporating multilingual support could improve future benchmarks.

\section{Ethical Considerations}

RICE-VL benchmarks cultural understanding in VLMs across Southeast Asia using culturally grounded tasks, with images and question-answer pairs annotated over 720 hour. Below, we outline key ethical challenges.

\paragraph{Annotator Involvement.} All annotators were recruited from Southeast Asia and underwent structured training to ensure high cultural fidelity. We acknowledge their contributions and the subjective judgments that may shape annotations.

\paragraph{Cultural Generalization and Representation.} Covering 11 ASEAN countries, the dataset may oversimplify minority, indigenous, and diaspora experiences. Future work should prioritize more nuanced cultural representations.

\paragraph{Stereotype Risk.} Some visual content may inadvertently reinforce cultural stereotypes. Although our intention was to capture authentic cultural elements, we recognize that the selection and framing of images might bias model perception. We implemented multiple layers of review to mitigate this, but residual bias may persist.

\paragraph{Content Bias and Privacy.} Some images may unintentionally reinforce cultural stereotypes despite efforts to capture authentic elements. Multiple review layers were implemented to mitigate bias, but some risk remains. The dataset
will undergo rigorous filtering to remove sensitive or identifiable content and will be released under an ethical use license, with documented filtering procedures to minimize harm.

\paragraph{Use of AI Tools.} ChatGPT was used only for early-stage grammar and fluency improvements. All core research tasks were independently conducted by the team.

RICE-VL aims to foster culturally inclusive VLMs. We urge the community to use it with cultural sensitivity and ethical commitment.

\section{Conclusion}

In this paper, we introduce RICE-VL, a culturally grounded benchmark designed to evaluate vision-language models across 11 Southeast Asian countries. RICE-VL includes over 28,000 human-curated question-answer pairs and 1,000 visual grounding annotations spanning 14 cultural categories, offering a high-resolution lens into cultural reasoning in multimodal systems. 

We evaluate six state-of-the-art VLMs across two tasks—culturalVQA and Visual Grounding, and observe significant disparities in performance between open-source and closed-source models. Additionally, performance varies across countries, with lower accuracy in underrepresented contexts such as Timor-Leste and Brunei.

Our results highlight the persistent limitations of current VLMs in handling culturally nuanced content, especially in low-resource settings. Prompt framing improves cultural localization, but deeper cultural reasoning remains a challenge. RICE-VL underscores the urgent need for culturally inclusive training data, evaluation strategies, and model design—paving the way toward equitable multimodal AI systems in the Global South.

\section*{Acknowledgments}

We would like to extend our sincere gratitude to the subject and language matter experts from the National Institute of Education (NIE), Nanyang Technological University (NTU), and the Singapore Institute of Technology (SIT) for their support and contributions throughout the research. This paper is supported by the National Research Foundation, Singapore under its AI Singapore Programme (AISG Award No: AISG2-GC-2022-004).

\bibliography{custom}

\clearpage

\appendix

\section{APPENDIX}

\subsection{RICE-VL Benchmark Categories}
\label{sec:appendix}

The RICE-VL benchmark is curated to assess the cultural understanding capabilities of Vision-Language Models (VLMs) within Southeast Asian contexts. It spans 11 countries—namely Singapore, Malaysia, Indonesia, Thailand, Vietnam, the Philippines, Cambodia, Laos, Myanmar, Brunei, and Timor-Leste—and captures 14 distinct cultural sub-ground categories. These categories were carefully selected to reflect the region’s rich socio-cultural, historical, and visual diversity.

Drawing inspiration from earlier cultural AI benchmarks, RICE-VL emphasizes culturally grounded content that extends beyond generic visual understanding. Each category encapsulates a unique aspect of Southeast Asian identity, shaped by centuries of tradition, belief systems, and community practices. Categories such as Architecture and Heritage, Festivals, Traditional Games, and Dance and Music celebrate the visual vibrancy of regional customs, while others like Marriage Customs, Religious Practices, and Clothing and Attire highlight deeply rooted, often localized expressions of culture.

In addition, RICE-VL includes visual representations of Food and Desserts, Drinks, Landmarks, Transport, Notable Key Figures, Painting, and Places of Worship. These domains were chosen not only for their cultural salience but also for their frequent appearance in public imagery and shared narratives across ASEAN societies. All data points were annotated by trained regional contributors and reviewed by cultural experts to ensure contextual fidelity.

Together, these cultural categories form the foundation for evaluating VLMs on tasks such as Visual Question Answering (VQA) and Visual Grounding (VG). Future versions of the benchmark aim to broaden the scope by incorporating folklore, oral traditions, and region-specific vernaculars, thereby enabling deeper cultural reasoning in multimodal AI systems.

\subsection{SEA-LAVE PROMPT}
% \label{sec:appendix}

To evaluate the cultural reasoning capabilities of Vision-Language Models in Southeast Asian contexts, we design SEA-LAVE (Southeast Asian Linguistic Agreement with Visual Evidence), a prompt-based evaluation framework that adapts the LAVE metric to culturally grounded tasks. Unlike traditional string-matching or generic semantic similarity metrics, SEA-LAVE incorporates region-specific cultural grounding by assessing answers across three dimensions: cultural relevance, sub-cultural insight, and country attribution. As shown in Table~\ref{tab:sea-lave_prompts}, we define tailored evaluation prompts for different task formats—open-ended Question-Answering, True/False statements, and Fill-in-the-Blank completions—ensuring consistency and interpretability across tasks. Each prompt instructs a model-as-judge to provide both discrete scores and qualitative justifications, allowing for fine-grained benchmarking of models’ cultural understanding.

As outlined in Table~\ref{tab:sea-lave_prompts}, each cultural task is guided by a prompt that emphasizes culturalVQA.

\begin{table*}[!htbp]
\centering
\renewcommand{\arraystretch}{1.4}
\begin{tabular}{|p{4cm}|p{9.5cm}|}
\hline
\textbf{Task Type} & \textbf{SEA-LAVE Prompt} \\
\hline
\textbf{1. Question-Answer Evaluation} &

YOUR TASK:
You are a cultural reasoning expert evaluating how accurately a model answers a cultural question based on Southeast Asian traditions.

Task: Assess the LLM's answer to the cultural question based on the following:

SCORING CRITERIA:
1. Answer Relevance (Binary: 0 or 1)
   - Score 1 if the answer reflects the correct sub-culture or aligns meaningfully with the question's cultural theme. Score 0 if unrelated.
2. Cultural Insight (Binary: 0 or 1)
   - Score 1 if the answer reflects cultural knowledge, such as symbols, practices, or traditions tied to the sub-culture. Score 0 if culturally generic or inaccurate.
3. Country Attribution (0, 1, or 2)
   - 0: No or incorrect country mentioned.
   - 1: Related country (e.g., another SEA country) but not the correct one.
   - 2: Correct country is mentioned.
4. Justification (Text)
   - Briefly explain the scores using specific elements from the answer. Note partial correctness where relevant.

EXAMPLES AND OUTPUT FORMAT:
Evaluate the following:
* Question: "{question}"
* Answer: "{llm response}"
* Culture: "{culture}"
* Sub-Culture: "{sub culture}"
* Country: "{country}"

Strictly output JSON:

\\
\hline
\textbf{2. True/False Statement Evaluation} &

YOUR TASK:
You are an expert verifying the cultural and geographic correctness of a True/False statement and its explanation provided by a language model.

Task: Analyze both the truth value and the explanation given by the model for cultural accuracy and alignment with the provided context.

IMPORTANT: Even though the answer is True or False, you are scoring the explanation using the following criteria:

SCORING CRITERIA:
1. Text Understanding (Binary: 0 or 1)
   - Score 1 if the explanation reflects the correct sub-culture or partially aligns with the cultural context. Score 0 if the explanation is off-topic or unrelated.
2. Cultural Understanding (Binary: 0 or 1)
   - Score 1 if the explanation includes any relevant cultural detail (e.g., practices, attire, foods, rituals). Score 0 if no relevant cultural context is present.
3. Country Score (Ternary: 0, 1, or 2)
   - 0: The explanation mentions the wrong or no country.
   - 1: The explanation mentions a related SEA country, but not the correct one.
   - 2: The correct country is mentioned, even if others are included.
4. Reasoning (Text)
   - Briefly justify each score using evidence from the explanation. Mention any partial correctness or mistakes.

EVALUATION CONTEXT:
* Statement (True/False Claim): "{llm response}"
* Culture: "{culture}"
* Sub-Culture: "{sub culture}"
* Country: "{country}"

Strictly output your evaluation in JSON format.

\\
\hline
\end{tabular}

\end{table*}

\begin{table*}[!htbp]
\centering
\renewcommand{\arraystretch}{1.4}
\begin{tabular}{|p{4cm}|p{9.5cm}|}
\hline
\textbf{Task Type} & \textbf{SEA-LAVE Prompt} \\
\hline

\textbf{3. Fill-in-the-Blank Evaluation} &
YOUR TASK:
As a cultural language expert, you are assessing the accuracy and appropriateness of a fill-in-the-blank completion about a cultural topic.

Task: Evaluate how well the model-filled phrase aligns with the cultural setting, terminology, and country of origin.

SCORING RUBRIC:
1. Phrase Appropriateness (0 or 1)
   - 1 if the completion is contextually correct and refers to the sub-culture. 0 if unrelated or inaccurate.
2. Cultural Relevance (0 or 1)
   - 1 if the phrase embeds cultural knowledge (e.g., rituals, foods, symbols). 0 if generic or missing cultural details.
3. Geographic Accuracy (0 to 2)
   - 0: Incorrect country.
   - 1: Related SEA country.
   - 2: Correct country mentioned or implied accurately.
4. Scoring Explanation (Text)
   - Describe in 2–3 sentences how the phrase reflects cultural and geographic accuracy.

EVALUATION SETUP:
* Prompt: "{question with blank}"
* LLM Response: "{llm response}"
* Culture: "{culture}"
* Sub-Culture: "{sub culture}"
* Country: "{country}"

Strictly return your evaluation in JSON format.

\\
\hline
\end{tabular}
\caption{Task-specific prompts used for cultural evaluation of model-generated responses under the SEA-LAVE framework.}
\label{tab:sea-lave_prompts}
\end{table*}

\subsection{Model Performance on CulturalVQA}
\label{sec:results_culturalvqa}

We evaluate the performance of various vision-language models on the CulturalVQA task, which involves answering culturally grounded questions based on visual input from Southeast Asian settings. 

\begin{figure*}[ht]
  \centering
  \includegraphics[width=0.8\linewidth]{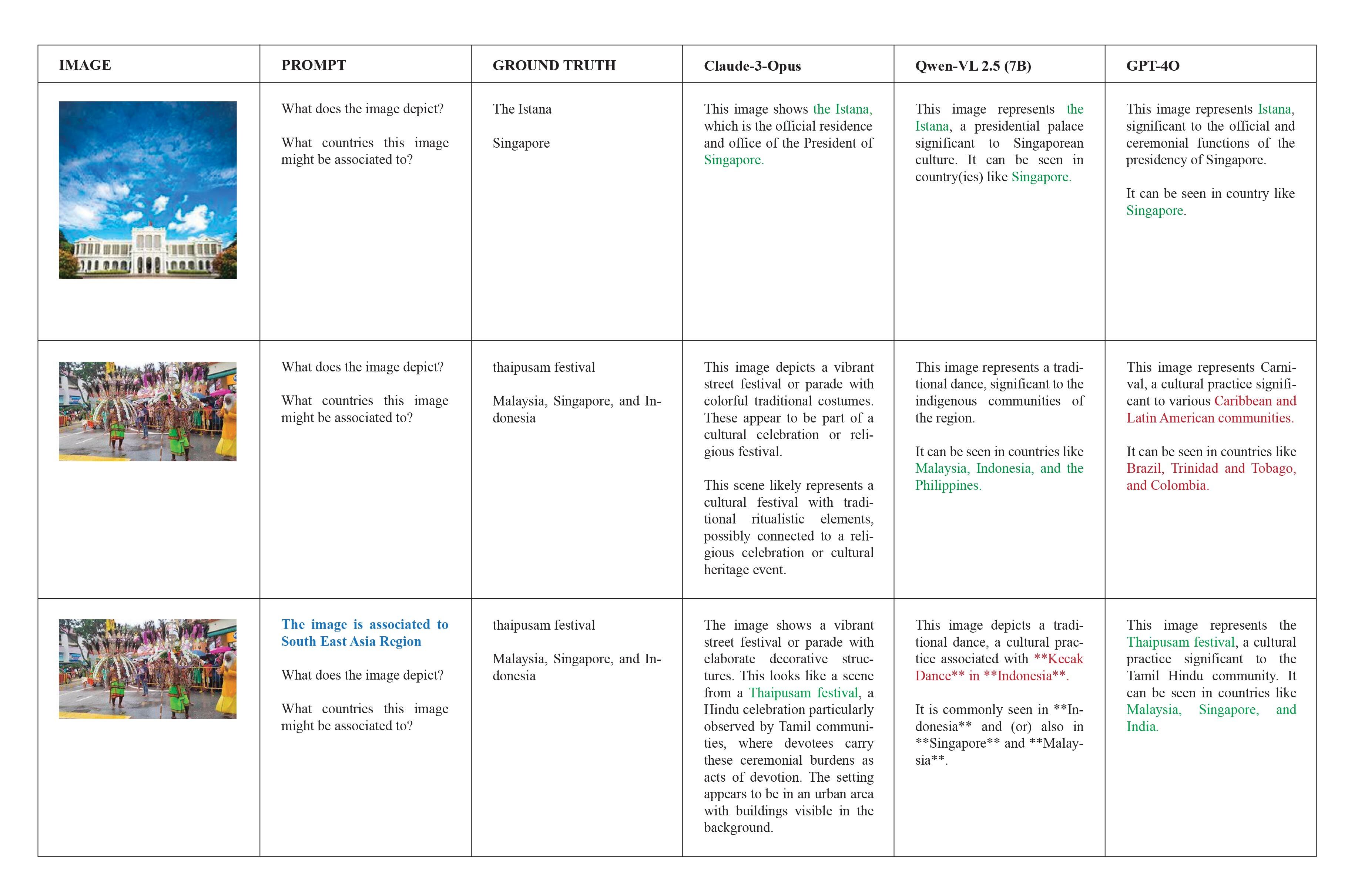}
  \caption{CulturalVQA results for cultural understanding of various models, global and SEA specific prompt. }
  \label{fig:task5}
\end{figure*}

As shown in Figure~\ref{fig:task5}, models such as GPT-4 and Claude-3 demonstrate superior performance on culturally nuanced queries, while smaller open-source models exhibit more variability across countries and question formats.

\subsection{Model Performance on Cultural Visual Grounding}
\label{sec:results_culturalvg}

We also assess the ability of models to localize culturally significant objects or scenes within images, captured under the Cultural Visual Grounding task.

\begin{figure*}[ht]
  \centering
  \includegraphics[width=0.8\linewidth]{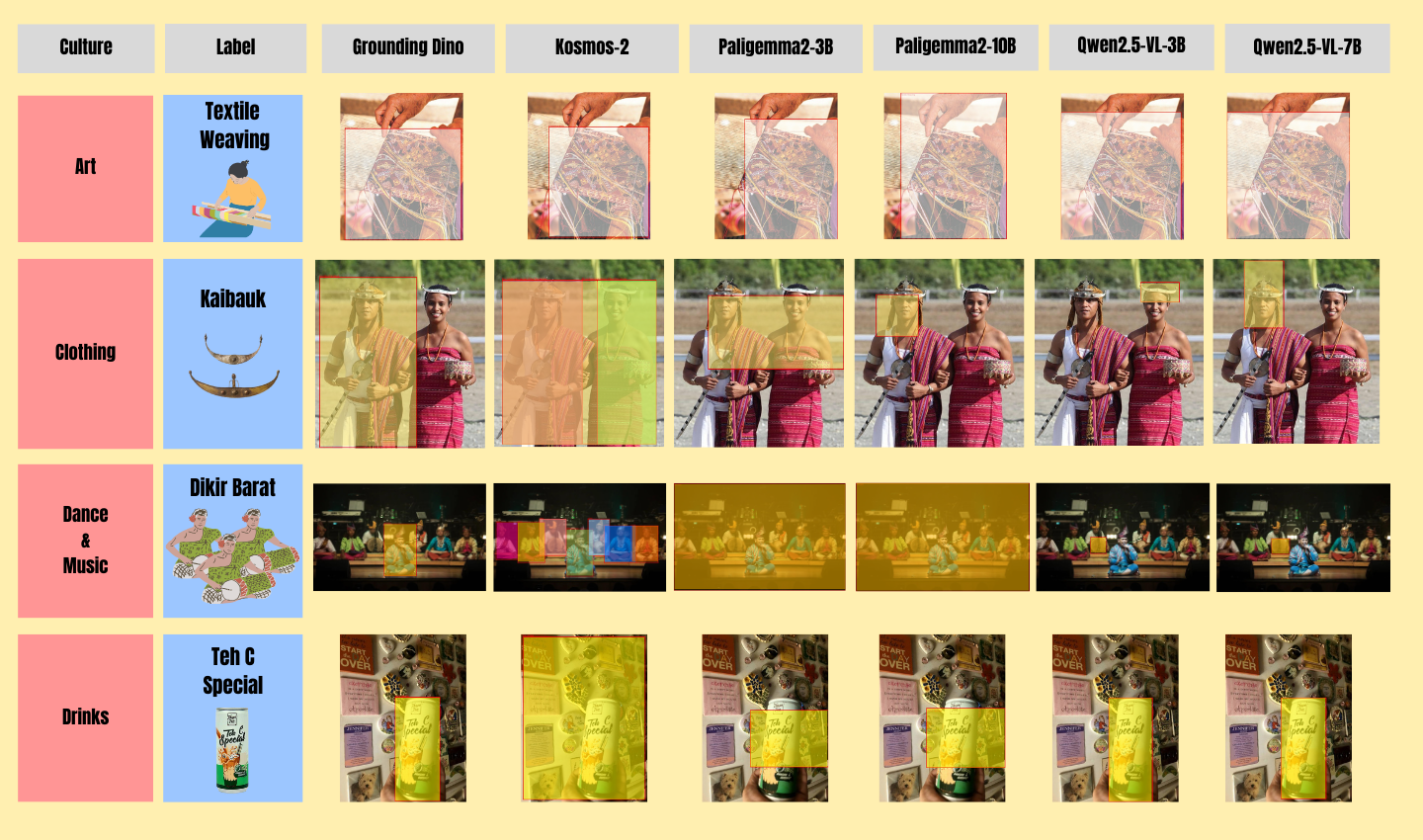}
  \caption{Visual Grounding results (Part 1): Comparing model predictions on region-specific cultural entities.}
  \label{fig:task10}
\end{figure*}

\begin{figure*}[ht]
  \centering
  \includegraphics[width=0.8\linewidth]{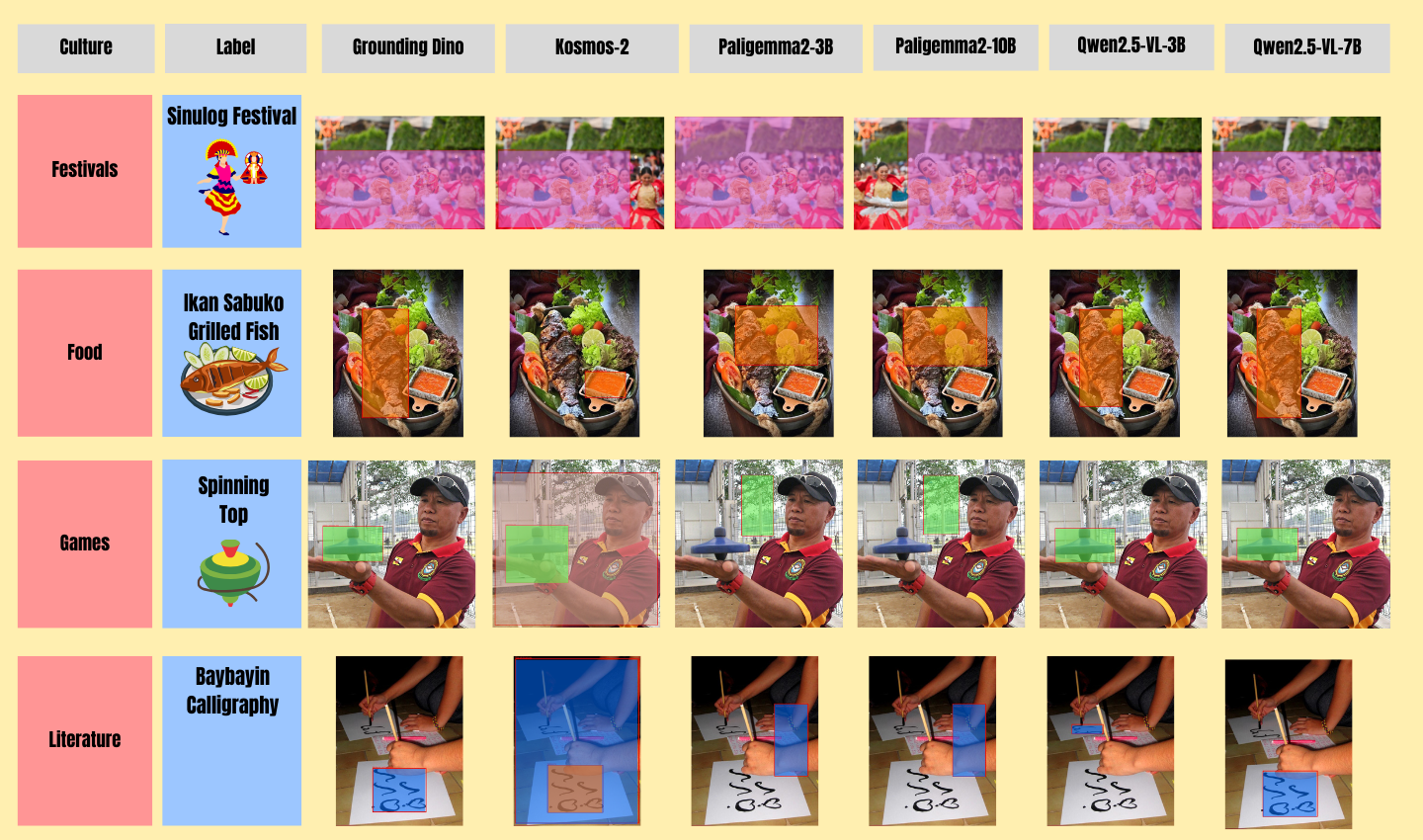}
  \caption{Visual Grounding results (Part 2): Comparing model predictions on region-specific cultural entities..}
  \label{fig:task11}
\end{figure*}

\begin{figure*}[ht]
  \centering
  \includegraphics[width=0.8\linewidth]{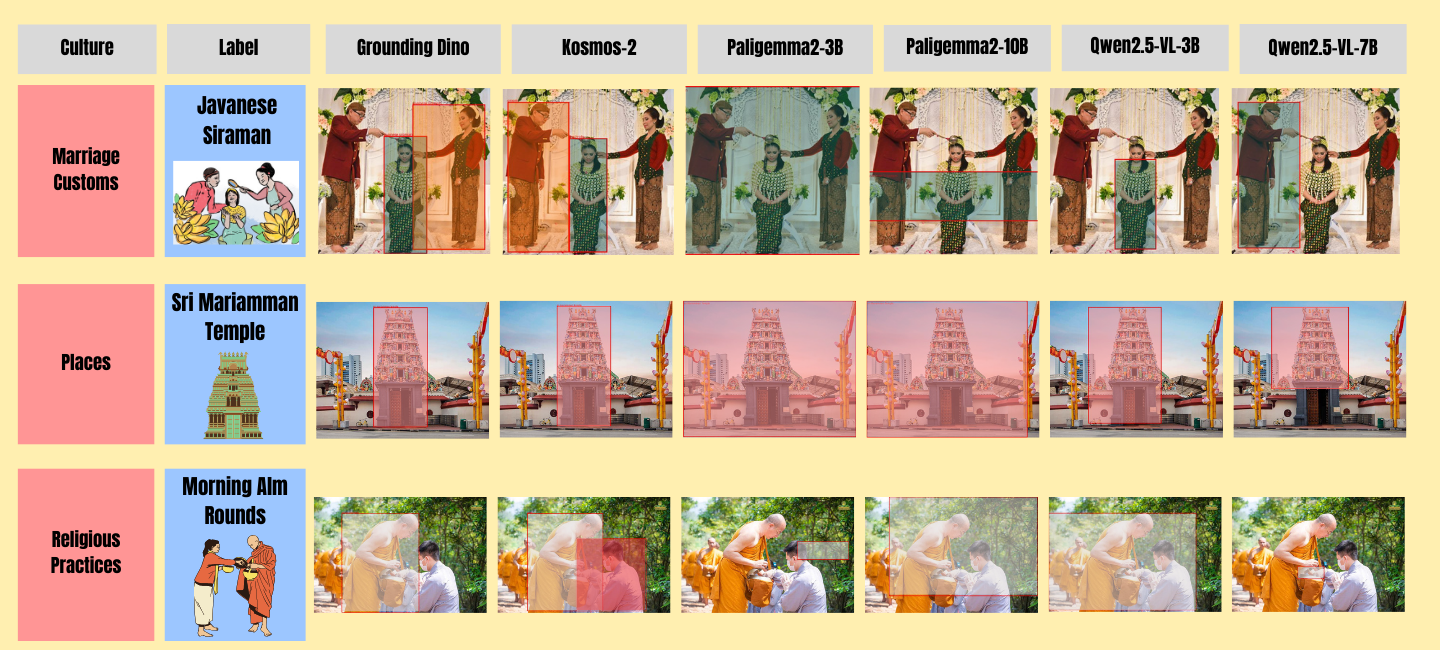}
  \caption{Visual Grounding results (Part 3): Comparing model predictions on region-specific cultural entities.}
  \label{fig:task12}
\end{figure*}

Figures~\ref{fig:task10}, \ref{fig:task11}, and \ref{fig:task12} illustrate qualitative comparisons across three representative categories. The visual grounding results reveal that models trained on culturally rich datasets are better at pinpointing region-specific artifacts such as traditional garments or religious structures, whereas general-purpose models often default to generic object detection.

% \subsection{Results of Various Models on culturalVQA}
% \label{sec:appendix}

% Figure~\ref{fig:task5} shows model performance on culturalVQA. 

% \begin{figure*}
%   \centering
%     \includegraphics[width=0.8\linewidth]{latex/images/eshan_table.jpg}
%   \caption{VG comparison.}
%   \label{fig:task5}
% \end{figure*}

% \subsection{Results of Various Models on cultural visual grounding}
% \label{sec:appendix}

% Figure~\ref{fig:task10}, ~\ref{fig:task11}, ~\ref{fig:task12} shows model performance on cultural visual grounding. 

% % 
% \begin{figure*}
%   \centering
%     \includegraphics[width=0.8\linewidth]{latex/images/1.png}
%   \caption{VG comparison.}
%   \label{fig:task10}
% \end{figure*}

% \begin{figure*}
%   \centering
%     \includegraphics[width=0.8\linewidth]{latex/images/2.png}
%   \caption{VG comparison.}
%   \label{fig:task11}
% \end{figure*}

% \begin{figure*}
%   \centering
%     \includegraphics[width=0.8\linewidth]{latex/images/3 (cropped).png}
%   \caption{VG comparison.}
%   \label{fig:task12}
% \end{figure*}

\end{document}